\documentclass[letterpaper]{article} 
\usepackage{aaai24}  
\usepackage{times}  
\usepackage{helvet}  
\usepackage{courier}  
\usepackage[hyphens]{url}  
\usepackage{graphicx} 
\usepackage{natbib}  
\usepackage{caption} 
\usepackage{algorithm}
\usepackage{algorithmic}
\usepackage{times}
\usepackage{epsfig}
\usepackage{graphicx}
\usepackage{amsmath}
\usepackage{amssymb}
\usepackage{booktabs}
\usepackage{multirow}
\usepackage{amssymb}
\usepackage{caption}
\usepackage{amsmath}
\usepackage{multirow}
\usepackage{subfigure}

\usepackage{newfloat}
\usepackage{listings}
\DeclareCaptionStyle{ruled}{labelfont=normalfont,labelsep=colon,strut=off} 
\floatstyle{ruled}
\newfloat{listing}{tb}{lst}{}
\floatname{listing}{Listing}
\title{Low-latency Space-time Supersampling for Real-time Rendering}
\author {
    Ruian He\equalcontrib,
    Shili Zhou\equalcontrib,
    Yuqi Sun,
    Ri Cheng,
    Weimin Tan\footnote{Corresponding authors: Bo Yan, Weimin Tan. This work is supported by NSFC (GrantNo.: U2001209 and 62372117) and Natural Science Foundation of Shanghai (21ZR1406600).},
    Bo Yan\footnotemark[2]
}
\affiliations {
    School of Computer Science, Shanghai Key Laboratory of Intelligent Information Processing, Fudan University\\
    rahe16@fudan.edu.cn, slzhou19@fudan.edu.cn, yqsun20@fudan.edu.cn, rcheng22@m.fudan.edu.cn, wmtan@fudan.edu.cn, byan@fudan.edu.cn
}

\usepackage{bibentry}

\begin{document}

\maketitle
\begin{abstract}
With the rise of real-time rendering and the evolution of display devices, there is a growing demand for post-processing methods that offer high-resolution content in a high frame rate. Existing techniques often suffer from quality and latency issues due to the disjointed treatment of frame supersampling and extrapolation. In this paper, we  recognize the shared context and mechanisms between frame supersampling and extrapolation, and present a novel framework, Space-time Supersampling (STSS). By integrating them into a unified framework, STSS can improve the overall quality with lower latency. To implement an efficient architecture, we treat the aliasing and warping holes unified as reshading regions and put forth two key components to compensate the regions, namely Random Reshading Masking (RRM) and Efficient Reshading Module (ERM). Extensive experiments demonstrate that our approach achieves superior visual fidelity compared to state-of-the-art (SOTA) methods. Notably, the performance is achieved within only 4ms, saving up to 75\% of time against the conventional two-stage pipeline that necessitates 17ms.
\end{abstract}


\section{Introduction}

Recently, the rendered contents have been a spotlight in Human-Computer Interaction (HCI) applications, spanning personal computers, mobile devices, and VR/AR hardware. Notably, the Meta Quest 2 VR headset supports an impressive 90 Hz frame rate with 1832x1920 resolution per eye. Meanwhile, Unreal Engine 5 can create photorealistic 3D environments using real-time ray tracing and physically-based shading. However, these advances pose a significant challenge for the limited hardware. Low resolution (LR) and low frame rate (LFR) significantly affect the user's immersion, as users are sensitive to both the details and the smoothness of rendered frames, which lies over 90Hz \cite{shi2023locomotion}. The post-processing methods provide a plausible solution to increase resolution and frame rate. 

\begin{figure}
  \centering
  \includegraphics[width=0.9\linewidth]{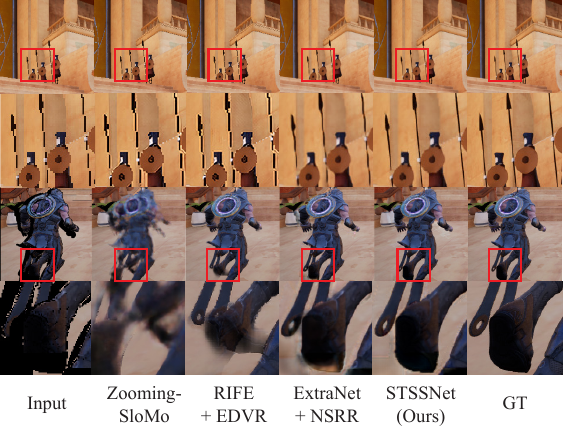}
  \caption{Visual comparison with the state-of-the-arts. The input has the aliasing  caused by low-resolution rendering (Up) and the warping hole by frame extrapolation (Down). Our method can address them well in a unified pipeline. }
  \label{fig:demo}
\end{figure}

\begin{figure}[t]
  \centering
  \includegraphics[width=\linewidth]{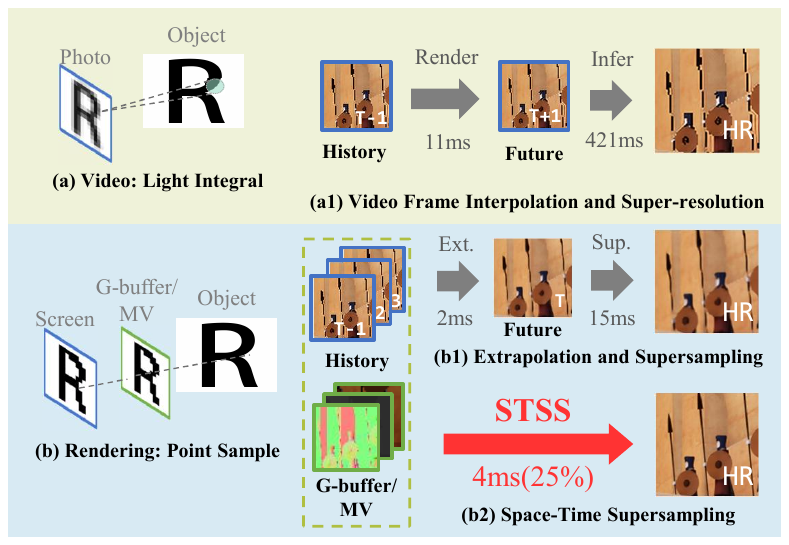}

  \caption{Comparison of different architectures in video and rendered content processing. We show the inference time targeting 1080P. Our framework can achieve better quality at a much faster inferencing speed.}
  \label{fig:teaser}
\end{figure}

Neural networks have demonstrated remarkable performance in enhancing video resolution and frame rate before their application to rendered content. Existing literature has explored video super-resolution \cite{liang2022vrt,li2023towards}, frame interpolation \cite{huang2022rife,plack2023frame} and space-time video super-resolution \cite{geng2022rstt,hu2023store}. Nonetheless, simply transferring these video processing methods to real-time rendering reveals two evident shortcomings, as shown in Fig.~\ref{fig:teaser}(a):  \textbf{(1) Not effective in addressing aliasing.} The aliasing in rendered content is caused by a low point sampling rate, while the pixels in photographic images are formed by integrating light rays. The fragmented object, such as the spear, can not be adequately compensated using conventional video methods. \textbf{(2) High processing latency.} The video methods primarily emphasize interpolation over extrapolation, which implies that the future frame ($T+1$) is needed first before generating the current frame ($T$). The methods unavoidably introduce extra display delay and spoil the user experience. Consequently, addressing these major drawbacks requires more effective methodologies tailored for real-time rendering.

Existing rendering techniques \cite{akeley1993reality,lottes2009fxaa,karis2014taa} enhance rendering quality by leveraging G-buffers and the motion vectors (MV). In real-time rendering, the engine first generates the G-buffers by fast preprocessing passes before the time-consuming frame rendering, as illustrated in Fig.~\ref{fig:teaser}(b). G-buffer contains the intrinsic details about objects before shading with light. We can generate base color, roughness, metallic, world normal, world position, scene depth, and many other results, covering from material to position. Also, the motion vectors (MV) can be calculated analytically from the 3D point displacements ahead of the rendering. MV facilitates warping and aligning the frames, akin to the role of optical flow in video processing. Consequently, G-buffer and motion vector provide practical information for post-processing. 

Recent state-of-the-art methods employ neural networks for learned rendering post-processing. Frame supersampling \cite{xiao2020neural} and frame extrapolation \cite{guo2021extranet} approaches are proposed to address the LR and LFR problems separately, as depicted in Fig.~\ref{fig:teaser}(b1). Though frame extrapolation has low latency, it introduces warping holes because of the occlusion. Meanwhile, supersampling methods still have problems with the aliasing. Previous methods fall short in terms of performance and efficiency without considering the shared context and mechanism in solving the aliasing and warping holes, as in Fig.~\ref{fig:demo}. 

Overall, the rendering post-processing calls for a unified approach to address the low resolution and low frame rate problem. Inspired by recent advancements in space-time video super-resolution(STVSR) works \cite{xiang2020zooming}, we introduce the concept of the space-time supersampling (STSS) problem. Specifically, a frame should be supersampled if it is rendered by the engine and, conversely, should be generated from previous frames if not rendered. Different from STVSR, STSS utilizes the G-buffer and MV provided by the rendering engines to meet the demand for higher resolution and frame rate (usually 1080P at 60Hz). Our contribution to the problem can be summarized as:

\begin{enumerate}
    \item To our best knowledge, we propose the first unified space-time supersampling (STSS) framework for real-time rendering, exploiting the common context and mechanism of frame extrapolation and supersampling. Our method achieves state-of-the-art performance among STSS pipelines. 
    \item Our framework streamlines the workflow by adopting a unified context and a shared neural network. It only requires 4ms per frame at 1080P settings, compared to 17ms by previous methods, saving up to 75\% time.
    \item We introduce a novel reshading mechanism that uniformly addresses the aliasing and warping holes as reshading regions. To achieve better reshading performance, we propose the Reshading Random Masking (RRM) and Efficient Reshading Module (ERM). 
\end{enumerate}

\begin{figure*}[t]
    \centering
  \includegraphics[width=0.83\linewidth]{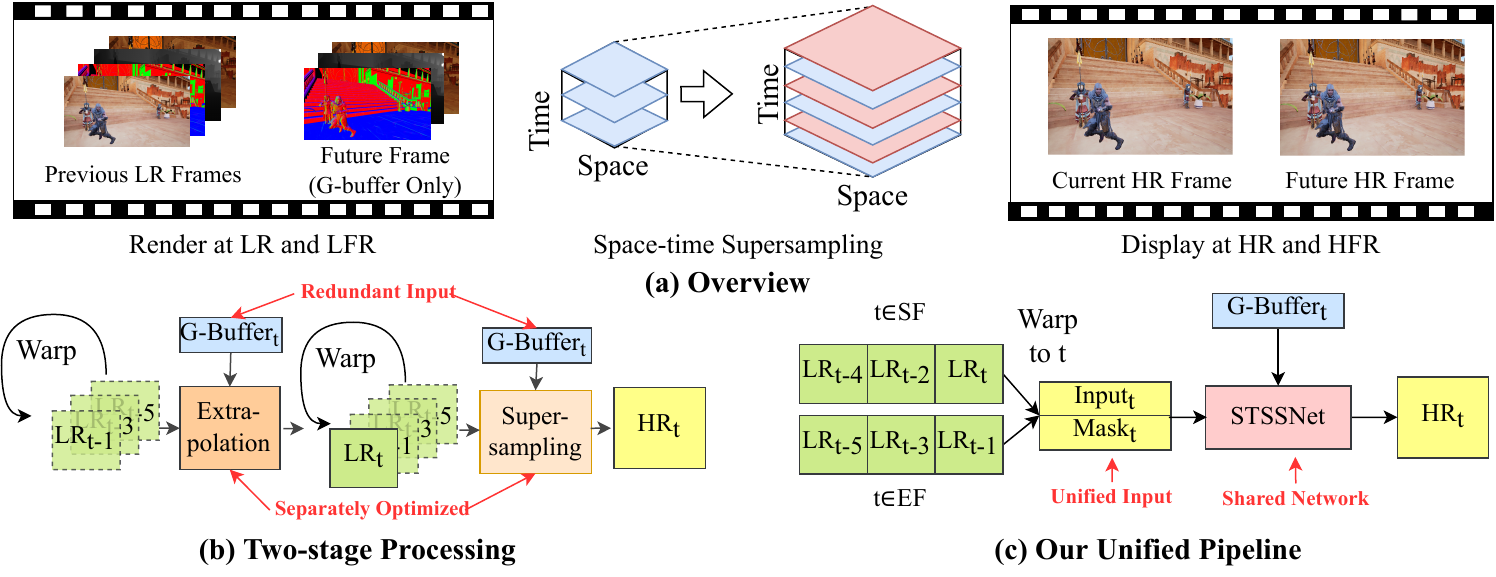}

  \caption{Overview of the space-time supersampling (STSS) pipeline. The pipeline generates the supersampled frame and the future frame, and users get HR results with a double frame rate. Previous works (b) separately execute frame extrapolation and supersampling with redundancy, while we exploit the common context and mechanism to build a unified pipeline (c).}
  \label{fig:flowchart}
\end{figure*}

\begin{figure}[t]
  \centering
  \includegraphics[width=0.9\linewidth]{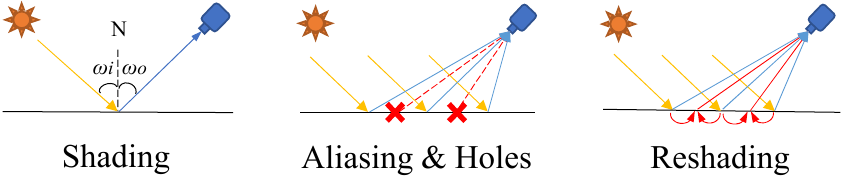}
  \caption{Illustration of shading and reshading mechanism. Aliasing and holes are not shaded in LR and LFR rendering.}
  \label{fig:reshade}
\end{figure}

\section{Related Work}

\subsection{Spatial and Temporal Supersampling}

The rendering content suffers aliasing because the content is undersampled in low resolution. One straightforward solution is supersampling antialiasing (SSAA), which downsamples the target frame rendered in a larger resolution. It will effectively reduce the aliasing but at a very high cost. Then, MSAA \cite{akeley1993reality}, MLAA \cite{reshetov2009morphological}, FXAA \cite{lottes2009fxaa} are proposed to improve the efficiency. Nevertheless, the spatial antialiasing methods ignore temporal coherence and may produce artifacts in large motion. TAA \cite{karis2014taa} is a widely-adopted algorithm that uses history frames to suppress flicker. TAAU \cite{epic2020taau} and FSR \cite{amd2022fsr} can further upsample the results while preserving consistency.

Recently, neural networks have been applied to real-time rendering. For supersampling, \cite{xiao2020neural,Thomas2020ARN,guo2022classifier} exploit the temporal feature to upscale the low-resolution frames and \cite{thomas2022temporally,li2022future} can also handle noisy LR renderings. For frame generation, \cite{briedis2021neural, Briedis2023KernelBasedFI} interpolates the offline rendered content, and \cite{guo2021extranet} is the first to extrapolate the future frame for a lower latency. There are also device-specific post-processing products for neural supersampling, such as DLSS \cite{edelsten2019dlss} and XeSS \cite{chowdhury2022xess}. DLSS3 \cite{henry2022dlss} further introduces a separate network for frame generation. Since most codes are unavailable, we compare with two open-sourced baselines via training, NSRR \cite{xiao2020neural} and ExtraNet \cite{guo2021extranet}, and compare other methods through provided software. Unlike previous works, our model has unified supersampling and extrapolation to achieve better quality and lower latency.

\subsection{Video Super-resolution and Frame Interpolation}

Video super-resolution (VSR) \cite{wang2019edvr,liang2022vrt,li2023towards} aims to increase the spatial resolution of the video. In contrast, video frame interpolation (VFI) \cite{ding2021cdfi,huang2022rife,plack2023frame} increases the temporal resolution, \textit{i.e.}, predicting the unknown frame between existing frames. Other than addressing the problems separately, the space-time video super-resolution (STVSR) \cite{xiang2020zooming,geng2022rstt,hu2023store} aims to expand the space and time resolution simultaneously, which is joint processing of VSR and VFI. We also notice existing video prediction and extrapolation methods \cite{oprea2020review,wu2022optimizing}. However, the video methods fail to consider the rendering intrinsic mechanism, leading to the aliasing problem and significant latency. 

\section{Methodology}

\subsection{Unified Space-time Supersampling Framework}
\label{sec:problem}

The low-resolution and low-frame-rate problems are pivotal concerns in real-time rendering and lead to poor user experience. Fig.~\ref{fig:flowchart}(a) illustrates the ideal prospect of rendering at LR and LFR and displaying at HR and HFR. Existing post-processing techniques have been proposed to address these issues individually. NSRR \cite{xiao2020neural} utilizes convolutional neural networks for spatial supersampling, while ExtraNet \cite{guo2021extranet} extrapolates future frames to achieve reduced display latency. 

An intuitive approach involves integrating current supersampling and extrapolation methods into a two-stage pipeline, as depicted in Figure~\ref{fig:flowchart}(b). However, this approach introduces significant redundancy in context and mechanism. The frame extrapolation and supersampling are executed independently with similar input of the previous frames and the G-buffer of the current frame. Then, both methods employ motion vectors to warp historical frames and leverage G-buffer for frame reconstruction. The only difference is the time step of input data. Consequently, the pipeline results in notable latency and poor performance.

\begin{figure*}[t]
  \centering
   \includegraphics[width=0.8\linewidth]{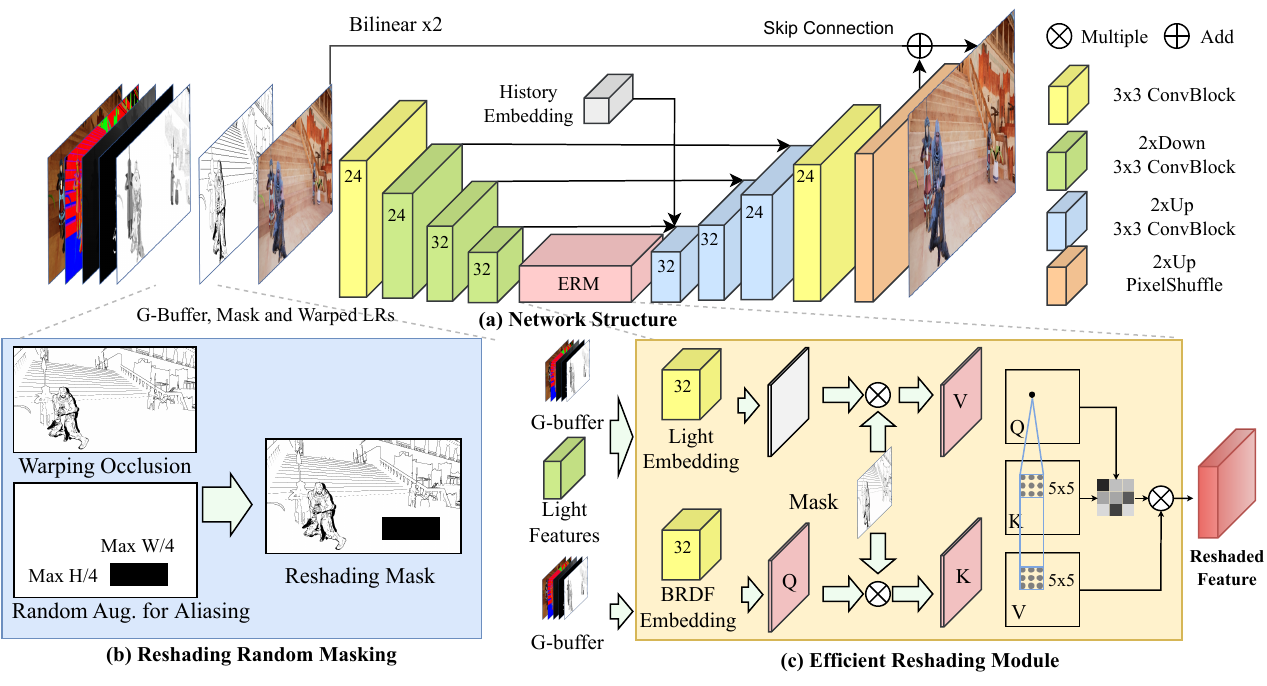}

   \caption{Network architecture of our unified space-time supersampling framework with RRM(b) and ERM(c). The numbers on the block denote the output channels of the convolution layer. 
   }
   \label{fig:framework}
\end{figure*}

In this paper, we first formulate our space-time supersampling(STSS) framework, as shown in Fig.~\ref{fig:flowchart}(c). Given the different time step $t$, the output $HR_{t}$ can be expressed as:
\begin{equation}
\label{eqa:strcuture}
HR_{t}=\left \{ \begin{matrix} 
\Phi(LR_{t-5,t-3,t-1},G_{t},MV_{t-5,...,t}), &t\in EF
\\
\Phi(LR_{t-4,t-2,t},G_{t},MV_{t-4,...,t}), &t\in SF
\end{matrix}  \right. 
\end{equation}
where $\Phi$ is the shared network for STSS. $G_{t}$ and $MV_{t}$ stand for G-buffer and MV at frame $t$. We denote the rendered frames as Supersampling Frames(SF) and the others as Extrapolation Frames (EF). We use three successive frames for SF, noted as $LR_{t}$, $LR_{t-2}$, and $LR_{t-4}$. Their G-buffer and MV are generated from $t$ to $t-4$ for every frame. The LR frames are all warped to $LR_{t}$ using the MV and producing warping masks. The warped LR frames, their warping mask, and the G-buffer are provided to the model to predict $HR_{t}$. On the other hand, we use $LR_{t-1}$ to $LR_{t-5}$ for EF. They are all warped to the future frame $LR_{t}$ using the motion vector from $t$ to $t-5$. STSS aims to minimize the perceptual difference between the generated and ground truth HR frames. So far, we unify the form to reduce the context redundancy.

\subsection{Reshading Mechanism}
\label{sec:problem}

The primary challenges of STSS comprise the aliasing resulting from low-resolution sampling and the warping holes due to frame extrapolation. Here, we dig deeper into the rendering principles for reducing the mechanism redundancy of addressing these two regions. The render engine performs shading (to calculate colors for every pixel) using the following equation\cite{kajiya1986rendering}:
\begin{equation}
\label{eqa:rendering}
    L_{o}(p, \omega_{o})=\int_{\Omega^{+}} L_{i}(p, \omega_{i}) f_{r}(p, \omega_{i}, \omega_{o}) \cos \theta_{i} \mathrm{d} \omega_{i}
\end{equation}
where, $L_{o}(p, \omega_{o})$ is the outgoing light at a shading point $p$ with direction $\omega_{o}$. $L_{i}$ is the incoming light from all possible directions, $\Omega^{+}$ and $\theta_{i}$ is the angle to the normal vector. $ f_{r}(p, \omega_{i}, \omega_{o})$ is the bidirectional reflectance distribution function (BRDF) describing the relation of incoming and outgoing radiance. BRDF and relative positions can be described with the G-buffer. We observe that a shared, significant characteristic between the aliasing regions and the warping holes is the absence of correct shading due to low rendering sampling rate in space and time. 

As shown in Fig.~\ref{fig:reshade}, we propose reshading mechanism (to re-calculate colors from shaded pixels) to treat these two regions unified as the reshading regions. The reshading mechanism involves a similar process as shading, which uses the BRDF and the light to calculate the reshaded feature, and an aggregation of adjacent regions. Specifically, we propose to replace the BRDF and the light in Eqa.~\ref{eqa:rendering} with G-buffer and predicted light features. Then, we reshade the color in the feature space, formulated as follows:
\begin{equation}
\label{eqa:content}
    \phi(p)=\int_{\Omega_p}(L(p+\delta)*G(p+\delta))R(p,p+\delta) \mathrm{d} \delta
\end{equation}
where $\phi(p)$ is the reshaded feature for the point $p$. $L$ is the light feature extracted from inputs, $G$ is the G-buffer and $*$ is a reshading function. We aggregate the reshaded feature in the adjacent region $\Omega_p$. $R(p,p+\delta)$ describes the similarity of $p$ and $p+\delta$, which serves as a normalization weight. Since the local shading depends on the BRDF and the visibility, we can infer $R(p,p+\delta)$ from the G-buffer and the mask. The reshading unifies the mechanism in the STSS problem and shows a feasible and efficient design.

\subsection{Network Architecture}

Fig.~\ref{fig:framework}(a) shows the structure of our proposed STSSNet. We adopt a U-Net \cite{ronneberger2015u} structure as our backbone. Our network takes as input the warped LRs (including $LR_{t-5,t-3,t-1}$), the G-buffer and the masks. The G-buffer includes base color, normal, depth, metallic, and roughness, which describes the BRDF and position in pixels. The input is first augmented with Reshading Random Masking (RRM). The mask is applied to the input and concatenated with the original one to feed into the network. Subsequently, we extract the features by several convolutional layers, followed by the Efficient Reshading Module (ERM). To capture temporal context, we utilize the history embedding\cite{guo2021extranet}, which encodes the previous frames and their warping mask. Then the features are fused with the history embedding and upsampled to get the final output. 

\paragraph{Random Reshading Masking.} To handle both the warping holes and the aliasing regions, we propose the concept of Random Reshading Masking (RRM), as depicted in Fig.~\ref{fig:framework}(b). Though previous applications of random masking have been observed in tasks like classification \cite{zhong2020random} and optical flow estimation \cite{teed2020raft}, RRM has a distinctive function. 

According to Eqa.~\ref{eqa:content}, the mask affects how the reshading mechanism aggregates from adjacent regions. But the input mask is limited to warping holes. Therefore, we use random augmentation to bring aliasing regions into the mask and make the model more robust. Concretely, we add randomly sized rectangles smaller than 1/4 of the frame's width and height at random positions to the occlusion mask. Then, we concatenate the modified LR images and G-buffers with the unaltered ones across channels for the network input. Also, we punish the masked areas with a larger loss.


\begin{figure*}[t]
  \centering
  \includegraphics[width=0.90\linewidth]{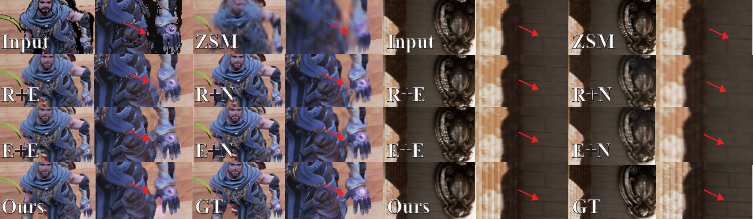}
   \caption{Qualitative comparison with SOTA methods. From left to right and up to down: Input, Zooming-SloMo, RIFE+EDVR, RIFE+NSRR, ExtraNet+EDVR, ExtraNet+NSRR, Ours, GT. The black regions are the warping holes and the red arrows are indicating the interested regions. Please zoom in for better view. }
   \label{fig:qualitative}
\end{figure*}

\begin{figure}[t]
\centering
\includegraphics[width=0.8\linewidth]{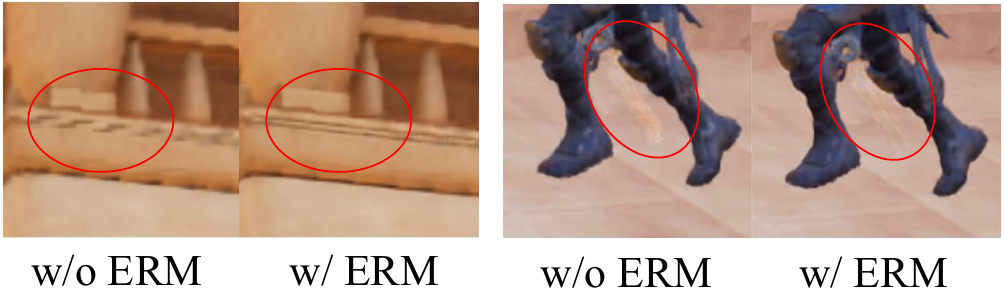}
\caption{Effect of ERM. The module can recover the details in both the aliasing regions and warping holes.}
\label{fig:ablation}
\end{figure}

\paragraph{Efficient Reshading Module.} Fig.~\ref{fig:framework}(c) shows the structure of ERM. While the backbone has partially addressed reshading problems through convolution layers, the awareness of G-buffer is lacking and the problems of the aliasing and warping holes still exist. To address the problems, our module incorporates a local attention mechanism to fuse features from nearby regions of the similar G-buffer.

Initially, we concatenate the downsampled G-buffer with the light features from the backbone, encoding them as the Light Embedding. Meanwhile, the G-buffer is encoded into the BRDF Embedding. Then, we take the BRDF Embedding as $\textbf{Q}$ map and multiply the Mask to the BRDF/Light Embedding as $\textbf{K}$/$\textbf{V}$ map for attention. $\textbf{Q}(p)$ is the attention query for every point $p$, including warping holes. We mask the value in the warping holes as zero, and get $\textbf{K}(p+\delta)$, which is the attention key for points $p+\delta$ in the local window $\Omega_p$ of the $p$:
\begin{equation}
\textbf{K}(p+\delta) =\left\{\begin{array}{ll}
\textbf{Q}(p+\delta) & \text { if } \mathbf{ValidMask}(p+\delta)=1 \\
0 & \text { otherwise }
\end{array}\right.. 
\end{equation}
It means that $\textbf{Q}(p) \textbf{K}^T(p+\delta)$ of warping holes $p+\delta$ in the local window is zero. Therefore, $\textbf{V}(p+\delta)$ of the warping holes (including the query point $p$ itself) will be ignored, and the only features from nearby valid regions are fused. 

Finally, the attention computation employs linear units \cite{zhang2021sparse} within a sliding window: 
\begin{equation}
\phi(p)=\sum_{\Omega_p} \operatorname{Relu}(\textbf{Q}(p) \textbf{K}^T(p+\delta)) \textbf{V}(p+\delta)
\end{equation}
where the attention is calculated only in the local $5\times5$ window $\Omega_p$ of each shading point $p$, so ERM has a small amount of computation (only 10\% of the backbone). Fig.~\ref{fig:ablation} shows that ERM gathers light embeddings from adjacent regions with similar BRDF and recovers the details.

\begin{table*}[t]
\centering
\scalebox{0.85}{
\begin{tabular}{@{}llcccccc@{}}
\toprule
  &  Scene   & RIFE+EDVR          & RIFE+NSRR         & ExtraNet+EDVR      & ExtraNet+NSRR     & ZoomingSloMo       & STSSNet(Ours)    \\ \midrule
\multirow{4}{*}{\rotatebox{90}{PSNR$\uparrow$}}  & LE & 30.38/29.57  & 34.28/32.28 & 30.45/30.29  & 34.29/33.87 & 30.39/28.51   & \textbf{35.02/34.72}   \\
                       & ST & 30.97/28.46 & 31.41/28.7  & 31.03/30.02  & 31.45/30.24 & 31.00/27.35      & \textbf{32.18/30.25}   \\
                       & SW & 28.32/26.05 & 29.93/27.01 & 28.71/28.32 & 30.18/29.88 & 28.17/24.79 & \textbf{31.00/30.35}   \\
                       & AR* & 31.52/28.95  & 33.26/29.59 & 31.52/31.42  & 33.14/31.97 & 31.36/25.86   & \textbf{33.36/32.40}   \\ \midrule
\multirow{4}{*}{\rotatebox{90}{SSIM$\uparrow$}}  & LE & 0.912/0.891  & 0.953/0.927 & 0.910/0.910  & 0.953/0.951 & 0.913/0.891   & \textbf{0.957/0.956}  \\
                       & ST & 0.918/0.862  & 0.921/0.864 & 0.920/0.909  & 0.923/0.911 & 0.920/0.851   &\textbf{ 0.926/0.913}  \\
                       & SW & 0.876/0.795  & 0.892/0.814 & 0.880/0.877   & 0.899/0.892 & 0.874/0.780   & \textbf{0.908/0.899}  \\
                       & AR* & 0.940/0.889  & 0.942/0.886 & 0.932/0.937  & 0.940/0.923 & 0.937/0.809   & \textbf{0.943/0.934}  \\ \midrule
\multirow{4}{*}{\rotatebox{90}{LPIPS$\downarrow$}} & LE & 0.135/0.131  & 0.090/0.097    & 0.132/0.135  & 0.080/0.089    & 0.050/0.054      & \textbf{0.018/0.020}   \\
                       & ST & 0.138/0.165  & 0.149/0.188 & 0.136/0.159  & 0.148/0.179 & 0.059/0.091      &\textbf{ 0.048/0.059} \\
                       & SW & 0.195/0.201  & 0.213/0.250 & 0.185/0.195  & 0.185/0.204 & 0.188/0.267   & \textbf{0.140/0.162}  \\
                       & AR* & 0.097/0.102    & 0.107/0.134 & 0.100/0.100   & 0.111/0.144 & 0.037/0.055      & \textbf{0.021/0.027}  \\ \midrule
\multirow{4}{*}{\rotatebox{90}{VMAF$\uparrow$}}  & LE & 56.33     & 69.66     & 59.38         & 73.41         & 59.25         & \textbf{78.85}   \\
                       & ST & 60.98     & 64.30      & 68.17         & 71.60          & 58.80          & \textbf{75.71}   \\
                       & SW & 54.79     & 64.47     & 64.48         & 76.87         & 49.98         & \textbf{81.16}   \\
                       & AR* & 72.60      & 78.84     & 80.37         & \textbf{86.05}         & 63.59         & 85.15   \\ \midrule
\end{tabular}
}
\caption{Quantitative comparison on PSNR, SSIM, LPIPS, VMAF. We show the indicators separately for the supersampling frames (SF)/ the extrapolated frames (EF). Lewis (LE), SunTemple (ST), Subway (SW), and Arena (AR) are the names of different scenes. The evaluation on AR (marked with *) follows the cross-scene setting. Bold text indicates the best results. 
}
\label{tab:quantitative}
\end{table*}

\paragraph{Loss Function.} Our loss function can be separated into two parts: weighted L1 loss and perceptual loss. First, we use the L1 loss to constrain the photometric consistency and punish the reshading regions for a larger weight. Our weighted L1 loss can be described as follows.
\begin{equation}
    L_{l1}  (I,\hat{I}) = \frac{1}{mn} \sum_{i=1}^{m} \sum_{j=1}^{n} w_{i,j} |I_{i,j} - \hat{I}_{i,j}|
\end{equation}
where $  (i,j)$ represents the point in the output $\hat{I}$ and the ground truth $I$. $m,n$ represents the height and width. $w_{i,j}$ is the loss weight, which is 2 in the reshading mask and 1 for other regions. Second, we add a perceptual loss \cite{zhang2018perceptual} to improve perceptual fidelity. This work adopts a pre-trained VGG \cite{simonyan2014very} network and calculates the per-layer similarity as the metric. Finally, our total loss function can be represented as $L(I,\hat{I}) = L_{l1}(I,\hat{I}) + w_{p}L_{p}(I,\hat{I})$, where $L_{p}$ is the perceptual loss, and $w_{p}$ is the weight. Empirically, we set $w_{i,j}=1$ and $w_{p}=0.01$.


\subsection{Datasets and Preprocessing}


In order to evaluate models for the novel STSS task, we propose a high-quality rendering dataset with LR-LFR and HR-HFR pairs. We combine the dataset construction protocol of the previous supersampling\cite{xiao2020neural} and extrapolation works \cite{guo2021extranet}.  \textbf{(1) Dataset Generation}: We construct scenes and actions, render them using Unreal Engine, and extract the G-buffer and motion vector. We turn on ray tracing and shadows for more realistic effects. It contains four scenes, Lewis(LE), SunTemple(ST), Subway(SW), and Arena(AR), each with 6000 frames for training and 1000 for testing. \textbf{(2) Resolution Configuration}: For LR-LFR frames, we turn off antialiasing and render at $960\times 540$ resolution at 30FPS. For HR-HFR frames, we first render at $3840\times2160$ resolution at 60FPS and then downsample it to $1920\times1080$ resolution as a $16\times$ supersampling ground truth. Therefore, supersampling frames (SF) take up 50\% of the frames and extrapolation frames (EF) for the other 50\%. \textbf{(3) Preprocessing}: The extrapolation motion vectors \cite{zeng2021temporally} are used to warp the frames. The motion vector, stencil, world position, normal, and NoV (dot product of the world normal and view vector) in G-buffer are used for generating warping masks following \cite{guo2021extranet}, while base color, metallic, roughness, depth, and normal for network input.

\begin{table}[t]
\centering
\scalebox{0.85}{
\begin{tabular}{@{}lcccc@{}}
\toprule
\multirow{2}{*}{Models} & \multicolumn{2}{c}{Edges} & \multicolumn{2}{c}{Warping Holes} \\ \cmidrule(l){2-5} 
                        & PSNR $\uparrow$           & SSIM  $\uparrow$          & PSNR $\uparrow$       & SSIM $\uparrow$       \\ \midrule
RIFE+EDVR               &  17.65               &  0.524                &  19.11           & 0.618            \\ 
ExtraNet+NSRR           &  23.36               & 0.853                &  25.40           &   0.849          \\
ZoomingSloMo            & 18.02                &  0.543               &  19.05           & 0.618            \\
\midrule
Ours w/o ERM            &  23.84               &  0.850                &   26.44           &   0.872          \\
Ours                    &  \textbf{23.95}      &  \textbf{0.853}      &  \textbf{26.52}           &  \textbf{0.874}           \\ \bottomrule
\end{tabular}
}
\caption{Comparison on different reshading regions.}
\label{tab:reshading}
\end{table}

\begin{table}[t]
\centering
\scalebox{0.85}{
\begin{tabular}{@{}lcccc@{}}
\toprule
\multirow{2}{*}{Models}  &  \multicolumn{2}{c}{Time(ms)$\downarrow$} & \multicolumn{2}{c}{Param(M)$\downarrow$}\\ \cmidrule{2-5}
& Stage1 &  Stage2 & Stage1 &  Stage2 \\ \midrule
RIFE+EDVR              & 19.2 & 402.2          & 9.8  & 20.7          \\
ExtraNet+NSRR          & 10.3(1.8) & 20.4(15.4) &  0.3 & 0.8          \\ \midrule
ZoomingSloMo           &  \multicolumn{2}{c}{821.1}            &  \multicolumn{2}{c}{11.1}              \\
STSSNet(Ours)          & \multicolumn{2}{c}{\textbf{9.0(4.4)}}       & \multicolumn{2}{c}{\textbf{0.4}}           \\ \bottomrule
\end{tabular}
}
\caption{Comparison of time cost and parameters. The bracket means a reimplementation with TensorRT.}
\label{tab:time1}
\end{table}

\begin{table}[t]
\centering
\scalebox{0.85}{
\begin{tabular}{@{}lccccc|c@{}}
\toprule
Size             & 1.LR    & 2.GB & 3.Warp       & 4.Network & Total & HR   \\ \midrule
720p       & 3.56  & 0.57     & 0.69       & 3.86  &   8.68 & 19.57  \\
1080p       & 5.69 & 0.63     & 0.81       & 4.35   &  11.51 & 38.38 \\
2160p       & 19.19 & 0.97     & 2.64       & 17.19  &   39.99 & 69.66 \\ \bottomrule
\end{tabular}
}
\caption{Time analysis on different resolutions in ms. }
\label{tab:time2}
\end{table}

\begin{table}[t]
\centering
\scalebox{0.85}{
\begin{tabular}{@{}llll@{}}
\toprule
Components               & Parameters & Computation & Time \\ \midrule
Backbone       & 141.00 K    & 20.53 GFLOPS      &  3.39 ms    \\
HistoryEmbedding & 120.70 K    & 8.78 GFLOPS      &  0.12 ms    \\
ERM           & 155.54 K    & 2.20 GFLOPS      &   0.83 ms   \\ \midrule
Full           & 417.24 K    & 31.50 GFLOPS      &   4.35 ms   \\ \bottomrule
\end{tabular}
}
\caption{Cost analysis for each component.}
\label{tab:computation}
\end{table}

\section{Experiments}

\subsection{Experimental Settings}

We use Pytorch to implement our model. We trained our model for 100 epochs on the training set with the Adam optimizer, and the learning rate was set to 1e-4. The learning rate optimizer is StepLR, with a step size of 50 and a gamma of 0.9. We use RandomCrop for augmentation, and each time we slice the input image into patches of size 256x256, repeat four times and feed them into the network. Following previous work \cite{guo2021extranet}, we evaluate for single-scene (train and test on LE, ST, and SW) and cross-scene (train on LE, ST, and SW, test on AR) settings. 

We compare various SOTA video and rendering methods in two folds. First, for two-stage methods, we combine four baseline models: RIFE \cite{huang2022rife} for VFI, EDVR \cite{wang2019edvr} for VSR, NSRR \cite{xiao2020neural} for rendering supersampling, and ExtraNet \cite{guo2021extranet} for rendering extrapolation. Next, we also compared the STVSR method ZoomingSloMo \cite{xiang2020zooming}. We trained the models for 100 epochs in single-scene setting and 50 epochs in cross-scene setting. We then tested them with an RTX 3090 GPU. The latest video methods \cite{zhou2021video,geng2022rstt,liang2022vrt} could not be tested because of the large GPU memory footprint over 24G.

\subsection{Comparison Against the State-of-the-arts}

\noindent\textbf{Space-time Supersampling.} Tab.~\ref{tab:quantitative} shows that our method has a significant advantage on PSNR, SSIM \cite{wang2004image}, LPIPS \cite{zhang2018perceptual}, and most VMAF \cite{li2018vmaf} indicators. LPIPS is widely used for perceptual image quality assessment, and VMAF is a famous perceptual video quality assessment algorithm developed by Netflix. Moreover, our method has more stable quality for both SFs and EFs, while other methods have a performance drop in EFs. The video processing methods have inferior performance, mainly because of ignoring the G-buffer and MV. 

\noindent\textbf{Reshading Regions.} Tab.~\ref{tab:reshading} demonstrates the effectiveness of our method on different reshading regions. We compare PSNR and SSIM metrics on the canny edge (thresholds set to 100 and 200) and the warping mask (interpolated from LR input) of EF. Fig.~\ref{fig:qualitative} also shows the qualitative comparison in the reshading regions on Lewis dataset. Our method can better handle edges and warping holes than previous methods.

\subsection{Efficiency Analysis}

\noindent\textbf{Inferencing Time Comparison.} Our tests run at 540P, targeting 1080P and on RTX 3090 GPU. All models run at half-precision for inference. We follow previous work \cite{xiao2020neural,guo2021extranet} to reimplement the rendering models using the TensorRT and show the results in the brackets. Tab.~\ref{tab:time1} shows that our method only has 0.4M parameters and needs 4.4ms per frame when inferencing, which saves 75\% of time for two-stage methods. In particular, the video methods need the future frame, further increasing the latency. 

\noindent\textbf{Time Analysis for Different Resolutions.} Our pipeline involves LR rendering, G-buffer generation, warping, and network inference. The LR rendering time is halved because we only render SF frames. Tab.~\ref{tab:time2} shows that we only need 30\% time compared to HR rendering of 1080P resolution. For instance, the Lewis scene runs at 26FPS for 1080P in the HR settings, and our method will increase the frame rate to 87FPS  ($3.3\times$) while preserving a high visual fidelity.

\noindent\textbf{Cost Analysis of Each Component.} Tab.~\ref{tab:computation} presents parameters, computational complexity, and runtime. The backbone network exhibits the highest computation, followed by the history embedding, while ERM has the lowest computation as it only operates features and performs locally.  

\begin{table}[t]
\centering
\scalebox{0.9}{
\begin{tabular}{@{}lccc@{}}
\toprule
Settings     & PSNR $\uparrow$            & SSIM $\uparrow$  \\ \midrule
Separately Optimized  &  34.39/34.29 & 0.952/0.951 \\
No G-buffer Input &  34.32/33.52 & 0.951/0.946 \\
\midrule
No RRM & 34.18/33.98 & 0.948/0.945  \\
No ERM & 34.91/34.62 & 0.956/0.954  \\
\midrule
STSSNet(Ours)    & \textbf{35.02/34.72} & \textbf{0.957/0.957} \\
\bottomrule
\end{tabular}
}
\caption{Ablation study on designs and modules.}
\label{tab:ablation}
\end{table}

\subsection{Ablation Study}

Tab.~\ref{tab:ablation} shows the ablation experiments on the Lewis scene. The results are evaluated separately for SF and EF using PSNR and SSIM. (1) To validate the unified framework, we independently trained our model on SF and EF. The performance dropped by 0.63dB because of not considering the common context and mechanism. (2) Removing G-buffer impacts performance by 0.7dB because it provides the BRDF information. (3) RRM has an average 0.7 dB impact because it greatly extends the reshading ability for both warping holes and the aliasing. (4) ERM can effectively exploit the similarity of BRDF and reshade the colors. It can improve an average of 0.1 dB on PSNR. Tab.~\ref{tab:reshading} also shows the effectiveness of ERM on edges and warping holes.

\subsection{Comparison Against Off-the-shelf Methods}

\begin{figure}[t]
\centering
  
\subfigure[Qualitative comparison.]{
  \includegraphics[width=0.9\linewidth]{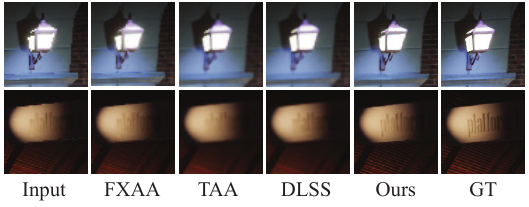}
}
   
\subfigure[User study results of preference.]{
  \includegraphics[width=0.9\linewidth]{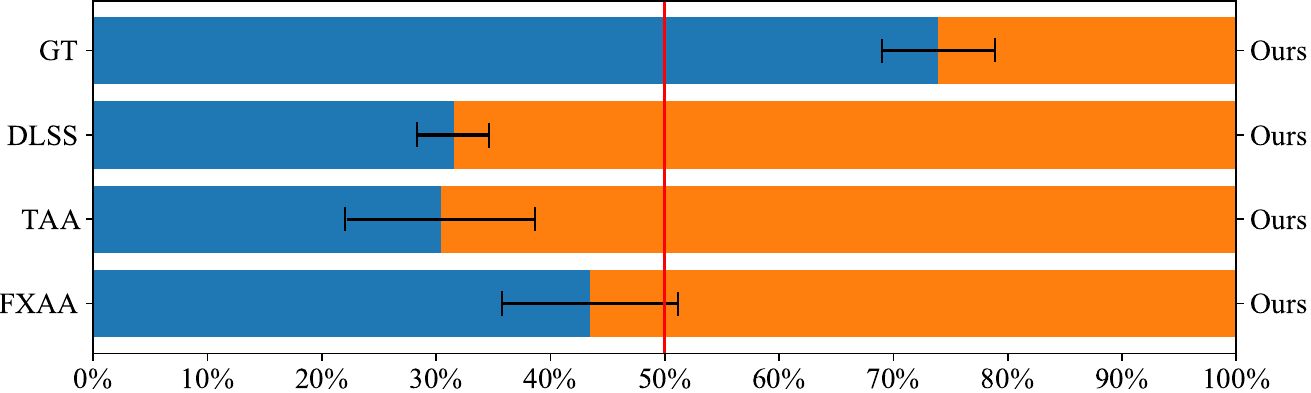}
}
\caption{Comparison against off-the-shelf methods.}
\label{fig:userstudy}
\end{figure}

Fig.~\ref{fig:userstudy}(a) shows the comparison with off-the-shelf super-sampling methods. We compare FXAA \cite{lottes2009fxaa}, TAA \cite{karis2014taa} and DLSS \cite{edelsten2019dlss}. They are optimized for different metric targets, so we compare them qualitatively. We use the rendering resolution of 540P for input and call them directly in the Unreal Engine through plugins. Our method can recover more details than previous methods while maintaining smooth edges. 

We also conduct a user study in Fig.~\ref{fig:userstudy}(b). 23 participants are provided with 32 random pairs of video sequences from the test set of the four scenes, with 250 frames each. Then, they are asked to choose the preferred one with higher rendering quality and less aliasing. We show the average preference and the variance. As a result, our method achieves higher preference than other methods. 

\subsubsection{Limitation and Discussion.} Our method may produce artifacts, and incorrect predictions for extreme motion and light conditions. Also, other techniques\cite{gpuopen2023fsr2, intro2020reflex}, offer many features to accelerate the rendering and displaying. Our method can be integrated with these techniques in the future to achieve better performance.

\section{Conclusion}

In this paper, we propose a novel unified Space-Time Super-Sampling (STSS) framework to improve the resolution and frame rate. We handle aliasing and occluded regions unified as reshading regions and propose Random Reshading Masking and Efficient Reshading Module. Extensive experiments demonstrate that our method can generate better results and inference much faster than previous methods. It can inspire future research on the STSS problem. 

\bibliography{aaai24}

\section{Appendix}

\subsection{Comparison on Single Tasks}

To demonstrate the benefit of a unified framework, we perform a conceptual experiment which ignores the pipeline and compares supersampling (Sup.) methods only on SFs and frame generation (Gen.) only on EFs. The frame generation methods are evaluated on the original 540P for a fair comparison. Tab.~\ref{fig:single} shows that we have an average 0.76dB PSNR advantage on each task.

\begin{table}[h]
\centering
\resizebox{0.9\columnwidth}{!}{%
\begin{tabular}{@{}lll|lll@{}}
\toprule
Sup.            & PSNR $\uparrow$          & SSIM  $\uparrow$         & Gen.           & PSNR $\uparrow$          & SSIM $\uparrow$   \\ \midrule
EDVR          & 30.45          & 0.911          & RIFE          & 32.21          & 0.931          \\
NSRR          & 34.29          & 0.953          & ExtraNet      & 33.88          & 0.950          \\
Ours & \textbf{35.02} & \textbf{0.957} & Ours & \textbf{34.67} & \textbf{0.955} \\ \bottomrule
\end{tabular}
}
\caption{Quantitative comparison on single tasks. }
\label{fig:single}
\end{table}

\subsection{Comparison with Two Stage Swapped}

In the main paper, we experiment with the assumption that the two stages of supersampling and interpolation/extrapolation have fixed order. The interpolation/extrapolation must be performed before supersampling. Therefore, we further swapped the two stages' order and showed the average metrics on the Lewis scene in Tab.~\ref{tab:misc}. The performance of two-stage methods is even more degraded than in the previous setting. One possible reason is that the HR G-buffers are unavailable in the second stage, and only the LR G-buffers can be used instead, which is unfavorable for extrapolation. The generated artifacts will accumulate when a frame is supersampled and then interpolated. 

\begin{table}[h]
\centering
\resizebox{0.7\columnwidth}{!}{%
\begin{tabular}{@{}lccc@{}}
\toprule
 Model         & PSNR $\uparrow$  & SSIM $\uparrow$   & LPIPS $\downarrow$  \\ \midrule
 EDVR+RIFE     & 29.66  & 0.8986 & 0.1349 \\
 NSRR+RIFE     & 33.25  & 0.9400    & 0.0847 \\
 EDVR+ExtraNet & 32.13 & 0.9297 & 0.1110  \\
 NSRR+ExtraNet & 34.05 & 0.9508  & 0.0848 \\
 STSSNet(Ours)          & \textbf{34.87}  & \textbf{0.9559}  & \textbf{0.0190}   \\ \bottomrule
\end{tabular}
}
\caption{Quantitative comparison of two-phase exchange. We swap the two stages to demonstrate the effect of stage order. 
}
\label{tab:misc}
\end{table}

\subsection{Comparison with Fine-tuned Models}

Since the video-based methods are usually pretrained on a larger dataset, we also compare the performance of finetuned models on Lewis in Tab.~\ref{tab:finetune}. We use the official models of RIFE, EDVR, and ZoomingSloMo as base models and finetune them for 100 epochs. Since NSRR and ExtraNet lack pre-trained models of large datasets, we train them from scratch to align with our method. Our method maintains the performance advantage.

\begin{table}[h]
\centering
\resizebox{0.9\columnwidth}{!}{%
\begin{tabular}{@{}lccc@{}}
\toprule
 Model         & PSNR $\uparrow$  & SSIM $\uparrow$   & LPIPS $\downarrow$  \\ \midrule
 RIFE+EDVR     & 34.47/32.73 &	0.949/0.923 &	0.023/0.032 \\
 RIFE+NSRR     & 34.14/32.35 & 	0.951/0.924 & 0.022/0.037 \\
 ExtraNet+EDVR & 34.49/34.30	& 0.949/0.947	& 0.023/0.024 \\
 ExtraNet+NSRR & 34.29/33.87 & 0.953/0.951 & 0.080/0.089 \\
 ZoomingSloMo  & 34.50/31.51 & 0.944/0.917 &  0.024/0.039 \\
 Ours          &\textbf{35.02/34.72} &\textbf{0.957/0.955}& \textbf{0.018/0.020} \\ \bottomrule
\end{tabular}
}
\caption{Quantitative comparison with fine-tuned video methods. }
\label{tab:finetune}
\end{table}

\begin{figure*}[t]
  \centering
  \includegraphics[width=0.98\linewidth]{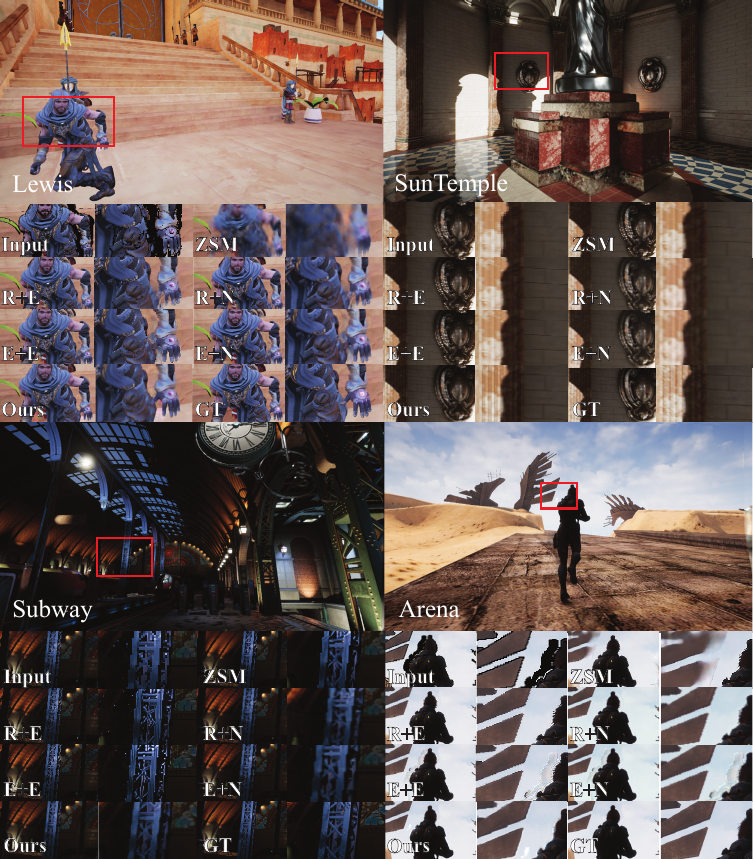}

   \caption{Qualitative comparison with SOTA methods. The first row is the example frames from the different scenes, and the following rows are cropped from the red triangle. From up to down and left to right: Input, Zooming-SloMo, RIFE+EDVR, RIFE+NSRR, ExtraNet+EDVR, ExtraNet+NSRR, Ours, GT. The black region in the input is the warping occlusion. Please zoom in for a better view. }
   \label{fig:qualitative}
\end{figure*}

\begin{figure*}[t]
  \centering
  \includegraphics[width=0.9\linewidth]{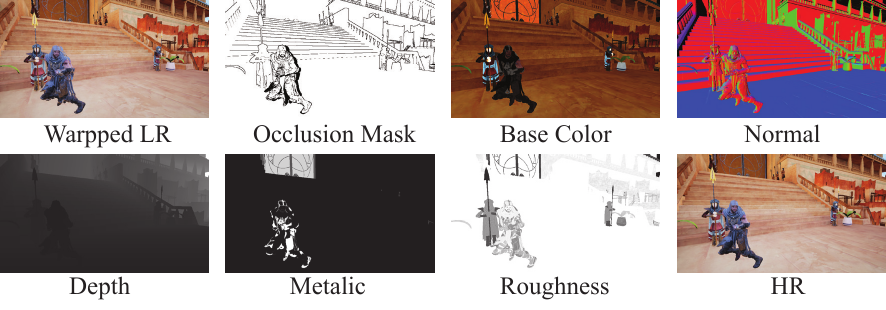}

   \caption{Visualization of the generated dataset. The G-buffer is composed of base color, normal, depth, metallic, and roughness. The warped LRs differ for extrapolating frames(EF) and supersampling frames(SF). EFs are warped in the future frame by motion vectors, but the SFs are warped to the present frame.}
   \label{fig:dataset}
\end{figure*}

\subsection{More Visualization}

Fig.~\ref{fig:qualitative} shows more qualitative comparison in Lewis, SunTemple, Subway, and Arena scenes. Our method achieves better quality in the details. 

Fig.~\ref{fig:dataset} show the visualization of the datasets. G-buffer for network input comprises base color, metallic, roughness, depth, and normal.


\end{document}